\documentclass[preprint,12pt]{elsarticle}

\usepackage{amssymb}
\usepackage{float}
\usepackage{hyperref}
\usepackage{graphicx}
\usepackage{subfigure}
\usepackage{epstopdf}
\usepackage{color}
\usepackage[noend]{algpseudocode}
\usepackage{algorithmicx,algorithm}
\usepackage{caption}
\usepackage{tabularx}
\usepackage{xcolor}
\usepackage{amssymb,amsmath}
\usepackage{pifont}
\usepackage{multirow}
\usepackage{amsmath}

\usepackage{hyperref}
\usepackage{caption}
\journal{Nuclear Physics B}
\hypersetup{
    colorlinks=true,            
    linkcolor=blue,              
    citecolor=blue,            
    filecolor=blue,   
    urlcolor=magenta            
}
\begin{document}
\captionsetup[figure]{labelfont={bf},name={Fig.},labelsep=period}
\captionsetup[table]{labelfont={bf}, labelsep=space }
\begin{frontmatter}



\title{Learning Implicit Neural Degradation Representation for Unpaired Image Dehazing}


\author[1]{Shuaibin Fan} 
\ead{fansb@stu.cqut.edu.cn}

\author[2]{Senming Zhong}
\ead{itcsmzhong@gpnu.edu.cn}

\author[1]{Wenchao Yan} 
\ead{yanwenchao@stu.cqut.edu.cn}

\author[1]{Minglong Xue\corref{cor}} 
    \ead{xueml@cqut.edu.cn} 

\cortext[cor]{Corresponding author.} 

\affiliation[1]{organization={College of Computer Science and Engineering},
            addressline={Chongqing University of Technology}, 
            city={Chongqing},
            country={China}}
\affiliation[2]{organization={Industrial Training Center},
            addressline={Guangdong Polytechnic Normal University}, 
            city={Guangzhou},
            country={China}}
\begin{abstract}
Image dehazing is an important task in the field of computer vision, aiming at restoring clear and detail-rich visual content from haze-affected images. However, when dealing with complex scenes, existing methods often struggle to strike a balance between fine-grained feature representation of inhomogeneous haze distribution and global consistency modeling. Furthermore, to better learn the common degenerate representation of haze in spatial variations, we propose an unsupervised dehaze method for implicit neural degradation  representation. Firstly, inspired by the Kolmogorov-Arnold representation theorem, we propose a mechanism combining the channel-independent and channel-dependent mechanisms, which efficiently enhances the ability to learn from nonlinear dependencies. which in turn achieves good visual perception in complex scenes. Moreover, we design an implicit neural representation to model haze degradation as a continuous function to eliminate redundant information and the dependence on explicit feature extraction and physical models. To further learn the implicit representation of the haze features, we also designed a dense residual enhancement module from it to eliminate redundant information. This achieves high-quality image restoration. Experimental results show that our method achieves competitive dehaze performance on various public and real-world datasets. This project code will be available at \href{https://github.com/Fan-pixel/NeDR-Dehaze}{https://github.com/Fan-pixel/NeDR-Dehaze}.
\end{abstract}



\begin{keyword}
Image Dehazing, Kolmogorov–Arnold, Implicit Neural Representation, Image Restoration


\end{keyword}

\end{frontmatter}



\section{Introduction}

Image dehazing is a highly challenging task in the field of computer vision, aiming to restore clear and detail-rich haze-free images from haze-contaminated inputs. This technique is crucial for enhancing the performance of downstream applications such as object detection (e.g., RT-DETR \cite{zhao2024detrs} and EEF \cite{hao2025eef}), autonomous driving (e.g., 
 VLP \cite{pan2024vlp}) and image segmentation (e.g., KACM \cite{xu2024kacm}, MDMASNet\cite{zhang2023mdmasnet} and FastICENet \cite{zhang2023fasticenet}). The formation of hazy images is commonly modelled via the Atmospheric Scattering Model (ASM):
\begin{eqnarray}
		&\boldsymbol I(x) = \boldsymbol J(x)t(x) + \boldsymbol A(1-t(x)),&
\end{eqnarray}
where $ \boldsymbol I(x) $ is the hazy image, $\boldsymbol J(x) $ is the scene radiance, $ t(x) $ is the transmittance, and $\boldsymbol A $ is the global atmospheric light. Usually defining $t(x)=\boldsymbol e^{-\beta d(x)}$. The $ \beta $ represents the atmospheric scattering coefficient, and $ d(x) $ is the scene depth.\par
\begin{figure}[h]
\centering
\includegraphics[width=0.5\linewidth]{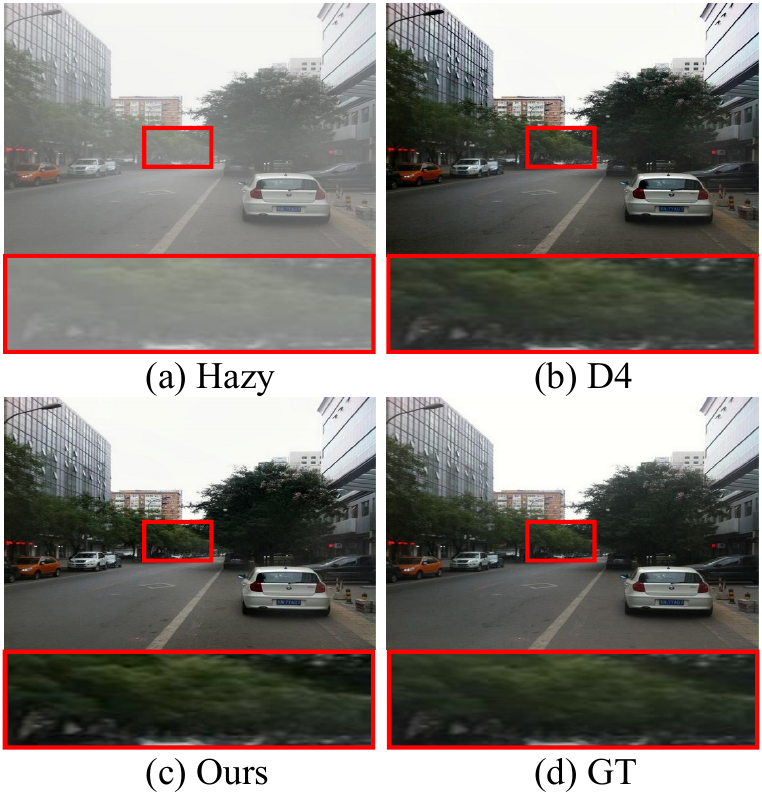}
\caption{A single image haze removal example. (a) The hazy image, (b) The dehazing result using the D4 method, (c) The dehazing result of our method, and (d) The ground-truth image.}
\label{fig:4_picture}

\end{figure}
Haze is extremely unevenly distributed in images and has complex multi-scale features \cite{ren2016single}, such as different sizes, shapes, angles, depths and densities. These characteristics pose a serious challenge to convolutional neural networks (CNNs) \cite{li2017aod}. Existing CNN architectures are difficult to effectively capture spatially varying features and non-local structural information due to fixed receptive fields, especially in images with uneven haze distribution. This limitation hinders their ability to effectively represent fine-grained features and achieve consistent global modeling, thus affecting the overall performance. Although haze images often exhibit similar patterns of visual degradation (e.g., typical haze-induced distortions), current methods rely heavily on traditional feature representation, which are sensitive to input variations and cannot fully model the implicit latent functions behind these common degradations. This limits their performance in complex scenes. Therefore, learning potential correlations between features from spatially varying haze has important research implications for recovering clearer images.\par
In this paper, we propose an unsupervised dehaze method called NeDR-Dehaze that focuses on implicit neural degradation representation. The method aims to deeply explore the nonlinear degradation features of haze and its implicit representation to recover clear and detail-rich image contents more efficiently. Specifically, to more effectively handle spatial feature variations and enhance the model's capacity for fine-grained feature representation and global consistency modeling, we propose a channel-independent and channel-dependent mechanism (KAN-CID). This architecture is based on the Kolmogorov-Arnold network (KAN) framework and integrates both channel-independent and channel-dependent mechanisms. Specifically, KAN is grounded in the Kolmogorov-Arnold representation theorem, which enables efficient decomposition of high-dimensional functions into multiple one-dimensional functions while adaptively adjusting the shapes of activation functions. This significantly improves the network’s representational capacity. Our proposed KAN-CID module balances local perception with global modeling, preserving structural details of the image and enhancing inter-channel contextual representation, thereby enabling more precise feature modeling.

In addition to learn the potential correlation between features from haze with spatially varying characteristics more efficiently and to enhance the adaptive representation of the model, we designed an implicit neural representation in the feature fusion part. On this basis, we further design the dense residual enhancement module (DREM). This module not only optimizes the processing of feature information but also greatly eliminates redundant information, which significantly enhances the model's ability to capture haze degradation features and spatial variation details. This improvement greatly enhances the robustness of the model in complex scenarios. As shown in Fig.~\ref{fig:4_picture}, our method shows significant improvement over current competitive methods. Our main contributions are summarized as follows:
\renewcommand{\labelitemi}{\scalebox{0.8}{$\bullet$}} 
\begin{itemize}
  \setlength{\itemsep}{0pt} 
  \setlength{\parsep}{0pt} 

\item To enhance fine-grained feature representation and global consistency modeling under non-uniform haze distribution, we propose a novel KAN-CID module, which integrates channel-independent and channel-dependent mechanisms within the Kolmogorov-Arnold network architecture.
\item We designed the implicit dense residual module to learn the haze degradation features and capture spatial details through continuous function mapping, thereby achieving more effective implicit haze modeling and stronger degradation processing capabilities.
\item We have conducted extensive experiments on several publicly synthesized and real-world hazed datasets, achieving impressive performance and demonstrating the robustness and effectiveness of our approach.
\end{itemize}
The remainder of this paper is organized as follows. In Section \ref{Related_work}, we provide a comprehensive review and summary of related work. Subsequently, Section \ref{Methodology} elaborates on the method we propose. In Section \ref{sec4}, we present the experimental results and analyze them in detail. Finally, Section \ref{Conclusion} summarizes the work presented in this paper.
\section{Related Work}
\label{Related_work}
\subsection{Prior-based Dehazing}
Traditional image dehazing methods primarily rely on prior knowledge to estimate haze thickness or transmission maps to restore clear images. A classical example is the Dark Channel Prior (DCP) \cite{he2010single}, which estimates the transmission map based on the statistical observation that local regions in haze-free images often contain pixels with low intensity. This method has achieved remarkable dehazing performance. Subsequently, approaches such as the Color Attenuation Prior (CAP) \cite{zhu2015fast} and Non-local Dehazing \cite{berman2016non} were proposed, which construct priors based on image brightness, saturation, or color distributions, thereby improving the adaptability and robustness of the algorithms. Although the a priori-based approach is physically interpretable, the high dependence of the model on predefined assumptions makes it prone to artifacts and color distortion when the input image deviates from these assumptions, which limits the processing effectiveness, especially in extreme weather or when the haze distribution is highly heterogeneous. 
\subsection{Learning-based Dehazing}
In recent years, deep learning has achieved breakthrough progress in the field of image restoration~\cite{xue2026ur2p, yan2026dfccnet, li2025curvpnp, chen2025rmanet}, making learning-based dehazing methods increasingly mainstream. Early approaches such as DehazeNet \cite{cai2016dehazenet} and AOD-Net \cite{li2017aod} leveraged convolutional neural networks to automatically learn the mapping from input hazy images to transmission maps or clear images, significantly improving dehazing performance. These methods typically require large amounts of synthetically generated hazy images paired with corresponding clear images as supervision.\par
To overcome the reliance on synthetic paired data, subsequent research has gradually shifted towards unsupervised or weakly supervised learning strategies. For example, Cycle-Dehaze \cite{engin2018cycle} employs a CycleGAN \cite{zhu2017unpaired} framework to achieve image-to-image dehazing without the need for paired training data. Meanwhile, methods such as FFA-Net \cite{qin2020ffa} and GridDehazeNet \cite{liu2019griddehazenet} further enhance model generalization by introducing attention mechanisms and multi-scale feature fusion modules.\par
Furthermore, to improve the practical applicability of dehazing models in real-world scenarios, recent research \cite{wu2023ridcp,wang2024ucl} has increasingly focused on evaluation using real hazy image datasets. SGDN \cite{fang2025guided} addresses the limitations of RGB-based representations by incorporating YCbCr structural information as guidance, leveraging dual-domain complementarity to better recover subtle edges and color relationships in real images. The RW2AH dataset was also introduced, offering rich geographic and climatic diversity, thus providing a more reliable foundation for training and evaluating dehazing models in real-world applications. ODCR \cite{wang2024odcr} proposes an orthogonal disentanglement contrastive regularization method, which enables more effective dehazing without paired data by orthogonally separating content and haze information and applying contrastive constraints. Lan et al. \cite{lan2025exploiting} proposed Diff-Dehazer, an unpaired image dehazing framework based on diffusion priors, which integrates physical priors with the generative capabilities of diffusion models to achieve superior dehazing performance on real-scene images. Existing methods often rely excessively on explicit feature extraction and physical models when facing complex scenes, resulting in difficulties in balancing between fine-grained feature capture of non-uniform haze distribution and global consistency modeling, which limits their dehazing effect and generalization ability in real-world scenes.
\subsection{Implicit Neural Representation} Implicit neural representation have been widely used in other fields \cite{chen2019learning,zhang2025learning} as well. For instance, Chen et al  \cite{chen2024bidirectional}. In the image deraining task, the implicit neural representation is integrated into the multiscale Transformer architecture to better explore multiscale information and model complex rain patterns to improve the robustness of the model in complex scenes. Nam et al. \cite{nam2022neural} proposed a framework utilizing Neural Image Representations (NIRs) to effectively merge multiple inputs into a single canonical view without the need to select one of the images as a reference frame. Yang et al \cite{yang2023implicit} proposed NeRCo, a collaborative shimmer image enhancement method based on implicit neural representation. It robustly recovers in an unsupervised manner the perceptually friendly results. However, local redundant features are encountered when processing images, which affects the modeling of high-frequency details and the ability to recover details in structured regions. Our work is inspired by the above field and further optimizes implicit representation to better facilitate the removal of haze.
	\section{Methodology}
	\label{Methodology}

\begin{figure}[h!]
    \centering
    \includegraphics[height=0.7\textwidth,width=1\textwidth]{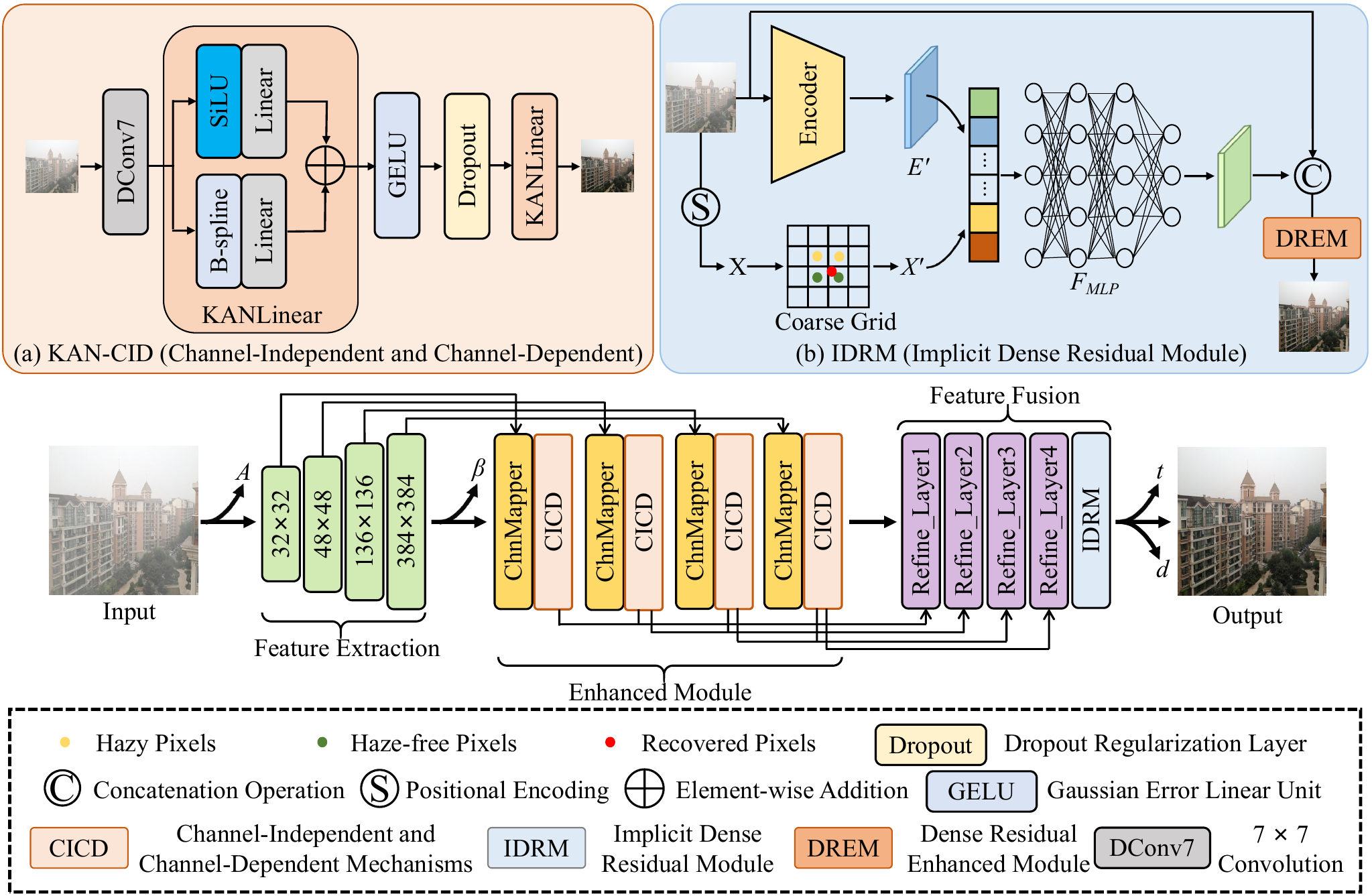}	
\caption{The architecture of the method consists of the dehazing network.  (a) represents the channel-independent and channel-dependent mechanisms based on the KAN network that we designed, and (b) denotes a dense residual enhancer based on Implicit Neural Representation. The feature extraction section denotes the extracted non-sized feature map; where $A$, $ \beta $, $t$, and $d$ are intermediate parameters of the atmospheric scattering model; the ChnMapper module converts feature maps with different numbers of channels into feature maps with a specific number of extended channels.}
\label{fig:frame}
\end{figure}

In our dehazing network architecture, after multi-scale feature extraction. First, the feature representation is further enhanced by KAN-CID module. Then, the image details are gradually restored under the multilevel feature fusion module. Finally, the implicit neural representation (IDRM) module is embedded in the image reconstruction process to model the haze degradation features as continuous functions using implicit neural representation to enhance the robustness and adaptability of the model. The redundant information in the high-frequency features is also removed to enhance the detail recovery of structural regions. The overall framework of our haze removal network is shown in Fig.~\ref{fig:frame}.\par
\subsection{KAN-CID Block}\label{subsec32}
In recent years, Kolmogorov-Arnold networks have received increasing attention from researchers \cite{li2025u,tran2025unpaired} as an emerging neural network architecture. Inspired by the Kolmogorov–Arnold representation theorem \cite{schmidt2021kolmogorov}, KAN aims to decompose high-dimensional functions into multiple one-dimensional functions, thereby enhancing the network’s representational capacity. Completely eliminating reliance on high-dimensional weight matrices enables the precise capture of arbitrarily complex dependencies among input variables with minimal parameters—a feat unattainable through traditional architectures like Transformers using high-dimensional self-attention or MLP. This flexible architecture offers potential advantages in image processing tasks. The theorem states that there exists a representation of the following form:
\begin{eqnarray}
		&f(x_1,x_2,\ldots,x_n)=\sum_{q=1}^{2n+1}h\left(\sum_{i=1}^{n}g_{q,i}(x_i)\right),&
\end{eqnarray}
where $ x = (x_1, x_2, \dots, x_n) $, it can be represented as a univariate continuous function $ h $ and a series of continuous bivariate functions $ x_i $ and $ g_q $, formed through composition. For any continuous function $f(x)$ defined in an $ n $-dimensional real space.\par
Our goal is to enhance the ability of dehazing networks to model complex haze distributions with improved fine-grained feature representation and global consistency, and we propose a feature enhancement module, KAN-CID, which integrates channel-independent and channel-dependent mechanisms. This module jointly models spatial and channel dimensions: the channel-independent branch extracts spatial details via spatial convolution, while the channel-dependent branch incorporates learnable kernel functions to capture inter-channel dependencies. This design enables a seamless fusion of local features and global contextual information.\par
In real-world image dehazing tasks, the non-uniform distribution of haze makes it challenging for convolutional neural networks to capture global semantics when relying solely on local feature extraction, often resulting in blurred details and color distortions. The KAN-CID module addresses this issue through its dual-branch architecture, significantly improving the network’s dehazing performance and image restoration quality.\par
As illustrated in Fig.~\ref{fig:frame} (a), the overall structure of the KAN-CID module consists of two branches: the Channel-Independent (CI) branch and the Channel-Dependent (CD) branch. In other words, the process initially conducts separate operations on each channel to deeply extract the information within the channels, and then models and integrates the relationships between channels through cross-channel information fusion. The implementation details are as follows: \par
\begin{eqnarray}
		& F_{CI}=DWConv_{7\times7}(F),&
\end{eqnarray}\par
\textbf{Channel-Independent Branch}. This branch performs depthwise separable convolution independently on each channel to extract spatial structural information. By avoiding inter-channel interference, it enables more precise preservation of local details such as edges and textures. Given an input feature map \( F \in \mathbb{R}^{C \times H \times W} \), the output of this branch is:\par
\textbf{Channel-Dependent Branch}. To accurately capture semantic correlations among channels, we design an innovative KAN-based network architecture. Specifically, the output from the Channel-Independent (CI) branch is first flattened and then fed into a series of carefully designed Channel-Dependent (CD) modules. These modules employ Kolmogorov–Arnold networks with kernels of varying sizes, enabling the efficient modeling of complex and multi-level semantic dependencies across channels. This design facilitates nonlinear cross-channel modeling. The process includes:
\begin{eqnarray}
		&KAN(I) = (\Phi_{N-1} \circ \Phi_{N-2} \circ \cdots \circ \Phi_1 \circ \Phi_0) I,&
\end{eqnarray}
where $\circ$ indicates element-wise multiplication, \( I \) denotes the input feature vector, and \( \Phi_i \) denotes the $i-th$ KAN-Layer.\par
Each $i-th$ KAN-Layer \( \Phi_i \), with \( n \)-dimensional input and \( n \)-dimensional output, can be represented as follows:
\begin{eqnarray}
		&\Phi = \{ \phi_{q,p} \}, p = 1, 2, \ldots, n_{in}, q = 1, 2, \ldots, n_{out},&
\end{eqnarray}
The Kolmogorov-Arnold representation theorem in matrix form can be expressed as:
\begin{eqnarray}
&\Phi_{\text{in}} = 
		\begin{bmatrix}
\phi_{1,1}(\cdot) & \cdots & \phi_{1,n}(\cdot) \\
\vdots & \ddots & \vdots \\
\phi_{2n+1,1}(\cdot) & \cdots & \phi_{2n+1,n}(\cdot) \\
\phi_{2n+2,1}(\cdot) & \cdots & \phi_{2n+2,n}(\cdot)
\end{bmatrix},&
\end{eqnarray}

\begin{eqnarray}
&\Phi_{\text{out}} = 
\begin{bmatrix}
\Phi_1(\cdot) & \cdots & \Phi_{2n}(\cdot) & \Phi_{2n+1}(\cdot)
\end{bmatrix},&
\end{eqnarray}
where $ \Phi $ includes $ n_{in} \times n_{out} $ learnable activation functions $ \phi $; $ \phi_{q,p} $ denotes the learnable parameters. \par
The computational results can be represented in the form of a matrix:
\begin{eqnarray}
&\Phi_i(I_i) = 
\begin{bmatrix}
\phi_{i,1,1}(\cdot) & \phi_{i,1,2}(\cdot) & \cdots & \phi_{i,1,n_i}(\cdot) \\
\phi_{i,2,1}(\cdot) & \phi_{i,2,2}(\cdot) & \cdots & \phi_{i,2,n_i}(\cdot) \\
\vdots & \vdots & \ddots & \vdots \\
\phi_{i,n_{i+1},1}(\cdot) & \phi_{i,n_{i+1},2}(\cdot) & \cdots & \phi_{i,n_{i+1},n_i}(\cdot)
\end{bmatrix}{I_i},&
\end{eqnarray}
where $ \Phi_i(I_i) $ denotes the output of feature $ I_i $ input to the $i-th$ KAN-Layer.\par
The design enhances the representation and global consistency of fine-grained features by combining channel-independent and channel-dependent mechanisms. Specifically, the channel-independent branch extracts spatial details through spatial convolution, while the channel-dependent branch captures inter-channel dependencies using a learnable kernel function. Seamless fusion of local features and global contextual information is achieved, which significantly improves the network's dehazing performance and image recovery quality.
\subsection{Implicit Dense Residual Module} In the field of image dehazing, most existing methods rely on explicit feature extraction and physical models. However, these approaches often exhibit limitations in complex real-world scenarios due to their inability to adapt to the diverse degradation characteristics of haze. To address this challenge, we propose an innovative approach to optimize implicit neural representation of hazy images in the form of dense residual enhancement. This method aims to efficiently restore image quality by learning implicit representation of haze degradation features. By leveraging the parameterization capabilities of neural networks, we represent the image content as a continuous function, enabling flexible capture of the intricate structures and variations within the image. Additionally, by performing a difference operation on the original input image, we effectively highlight the key image contents that are obscured by haze. \par
\begin{figure}
\centering
\includegraphics[width=\linewidth]{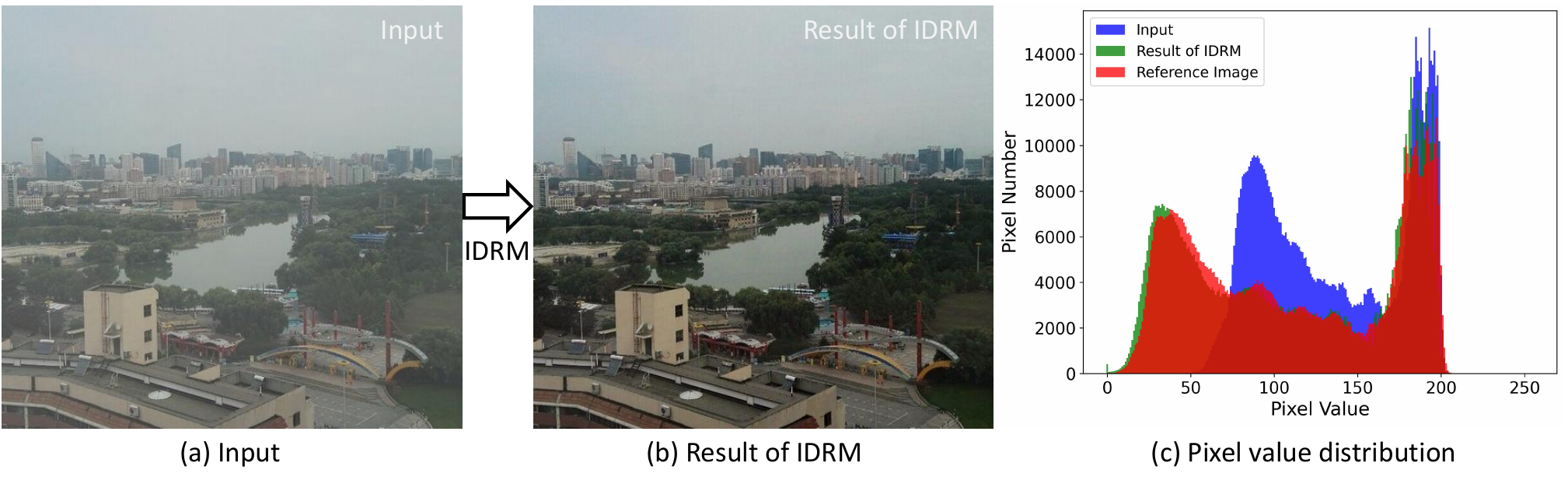}
\caption{Visual comparison between the input image and the IDRM result. It can be clearly observed that IDRM has a significant effect on reducing the high-intensity pixel values caused by haze. This method effectively mitigates the interference of haze in the image and reconstructs the underlying haze-free image, thereby significantly enhancing the clarity and detail representation of the image.}
\label{fig:idrm}
\end{figure}
As shown in Fig.~\ref{fig:frame}, in the feature fusion section, an improved feature map  \( I_{\text{refined}} \in \mathbb{R}^{H \times W \times 3} \) is obtained through a four-step refinement layer. Where \( H \times W \) denotes the spatial resolution of the image, we transform the hazy image into a feature map \( E' \in \mathbb{R}^{H \times W \times C} \). Meanwhile, the corresponding 2D coordinates are recorded in \( X' \in \mathbb{R}^{H \times W \times 2} \). We then fuse \( E' \) and $ X' $, and pass the combined information through a multilayer perceptron ($F$) for decoding to generate the final neural representation.
\begin{eqnarray}
		&I_{\text{enhanced}}[i,j] = F(E'[i,j], X'[i,j]),&
\end{eqnarray}
where $ I_{\text{enhanced}}[i,j] $ represents represents the processed pixel value at position $(i,j)$,  $ E'[i,j] $  is the feature vector at that position, and $ X'[i,j] $ is the corresponding coordinate vector. \par
To eliminate redundant features and capture common haze degradation characteristics, we integrate into the dense residual enhanced module and train a multilayer perceptron (MLP) by learning the following mapping function:
\begin{eqnarray}
&f_\theta : \mathbb{R}^2 \rightarrow \mathbb{R}^3, \, f(x, y) = (r, g, b),&
\end{eqnarray}
where \(f_{\theta} : \mathbb{R}^2 \rightarrow \mathbb{R}^3\) represents a mapping defined by parameters \(\theta\), which projects two-dimensional image coordinates into a three-dimensional color space, where \((r, g, b)\) denote the values of the red, green, and blue color channels, respectively.\par
To enhance the network's ability to capture spatial information, we introduce a spatial encoding mechanism. Specifically, we map the image coordinates $ \textit{X} $ through an encoding function $ \gamma$ into a higher-dimensional space, which is then used as one of the inputs to the network. This process can be represented by the following formula:
\begin{eqnarray}
&\textit{X}' = \gamma(\textit{X}), &
\end{eqnarray}
\begin{flalign}\label{eq7}
\gamma(\textit{X}) = \left[ \sin(2^0 \pi \textit{X}), \cos(2^0 \pi \mathbf{X}), \ldots, \right. & \nonumber \\
\left. \sin\left(2^{L-1} \pi \textit{X}\right), \cos\left(2^{L-1} \pi \textit{X}\right) \right], &
\end{flalign}
where \( \textit{X} \) represents the image coordinates, \( \textit{X}' \) denotes the encoded coordinates, and \( L=4 \) is the hyperparameter that determines the dimensionality of the encoding. \par
\begin{figure}
\centering
\includegraphics[width=\linewidth]{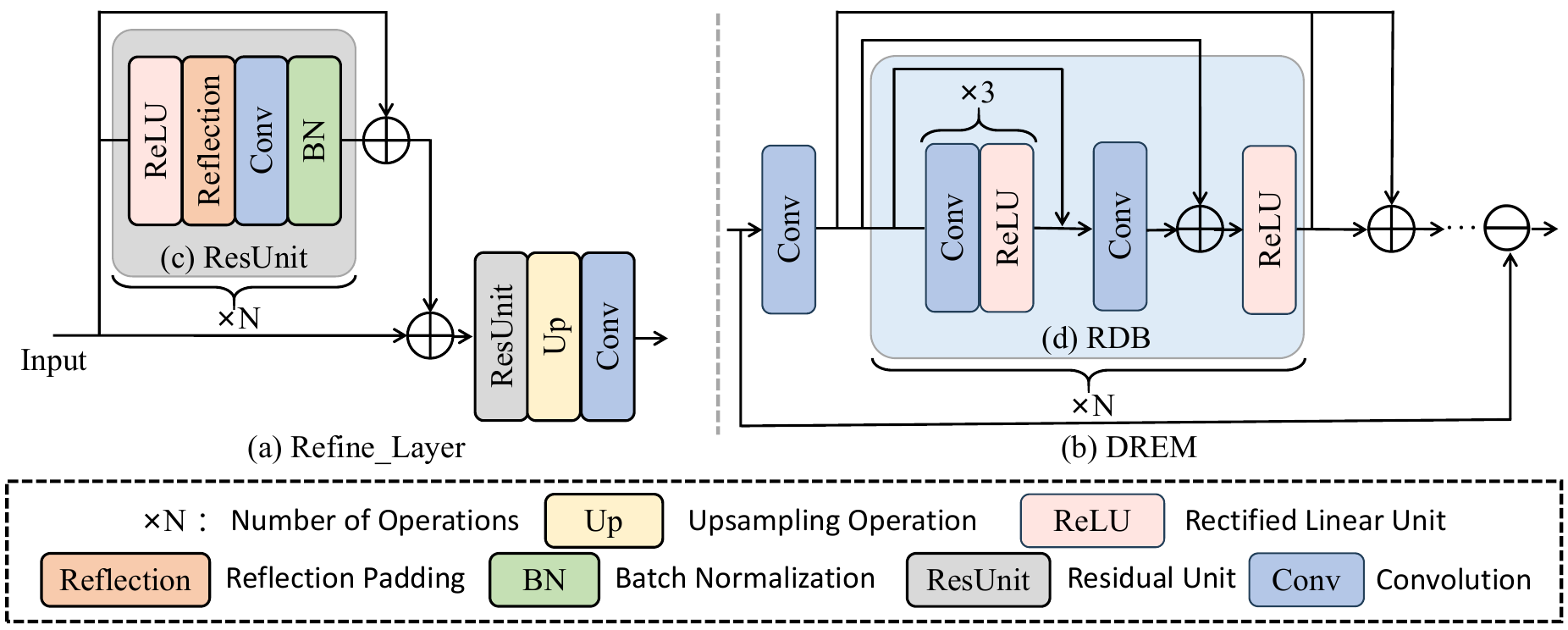}
\caption{The detail branch of the dehazing network. (a) Feature Fusion Module, (b) Dense Residual Enhanced Module, (c) Residual Convolutional Module, and (d) Residual Dense Block.}
\label{fig:frame2}
\end{figure}

Finally, as shown in Fig.~\ref{fig:frame2} (a), after coarse fusion of features of different sizes, our designed implicit dense residual module (IDRM) performs deep fusion refinement, with the specific structure shown in Fig.~\ref{fig:frame} (b). This module is constructed based on implicit neural representation. Unlike existing implicit neural network methods, we effectively remove redundant features from high-frequency features and enhance the detail recovery capability of structural areas through the dense residual enhanced module. Detailed content is introduced in Section~\ref{DREM}, and it also provides a novel and efficient paradigm for modeling image content. \par
Compared to traditional explicit representation methods, IDRM demonstrates superior flexibility in capturing complex structures and subtle variations within images, making it particularly well-suited to address the diverse and nonlinear degradations caused by haze. Specifically, the incorporation of local aggregation and feature unfolding mechanisms significantly enhances the network’s capability to model fine-grained local features and texture information. In addition, the use of positional encoding further enriches the network’s spatial awareness, enabling a deeper understanding of scene geometry. Notably, the IDRM adopts an end-to-end training strategy, allowing the model to directly learn optimal feature representation from raw input images without relying on handcrafted priors or physical models. This not only simplifies the dehazing pipeline but also improves both performance and efficiency. As shown in Fig.~\ref{fig:idrm}, the IDRM exhibits significant advantages in image dehazing, effectively suppressing high-intensity white haze and restoring a clear image.

\subsection{Dense Residual Enhanced Module}
\label{DREM}
In the framework described above, we further design a dense residual enhanced module (DREM) to enhance the expressive capability of IDRM in detailed modeling. The IDRM structure is capable of effectively learning the continuous distribution of the overall image structure, but it inevitably encounters some locally redundant features during the processing. To address this, DREM is embedded as an auxiliary module during the training process, focusing on removing redundant information from high-frequency features and enhancing the detail recovery ability in structural regions.\par
As shown in Fig.~\ref{fig:frame2} (b), the DREM module integrates residual learning and dense connectivity mechanisms to construct a double-layer residual dense block (RDB) structure. Through the initial convolutional layer and residual paths, feature responses are reinforced, and then dense connections and local fusion progressively enrich the feature hierarchy. The final output features are refined through convolution and, by performing a differential operation with the original input image, effectively highlight key image content under hazy interference, assisting IDRM in more accurately fitting the implicit representation of the target image. This design improves the network's adaptability to complex hazy distributions without relying on traditional explicit estimation, while also compensating for IDRM's structural expression capability in high-frequency detail modeling, providing stronger restoration ability for the image dehazing task.

\section{Experiments}\label{sec4}
In this section, to evaluate the performance of the proposed method on the dehazing network, we conducted a series of qualitative and quantitative comparison experiments.
\begin{figure}[h]
\centering
\includegraphics[width=\linewidth]{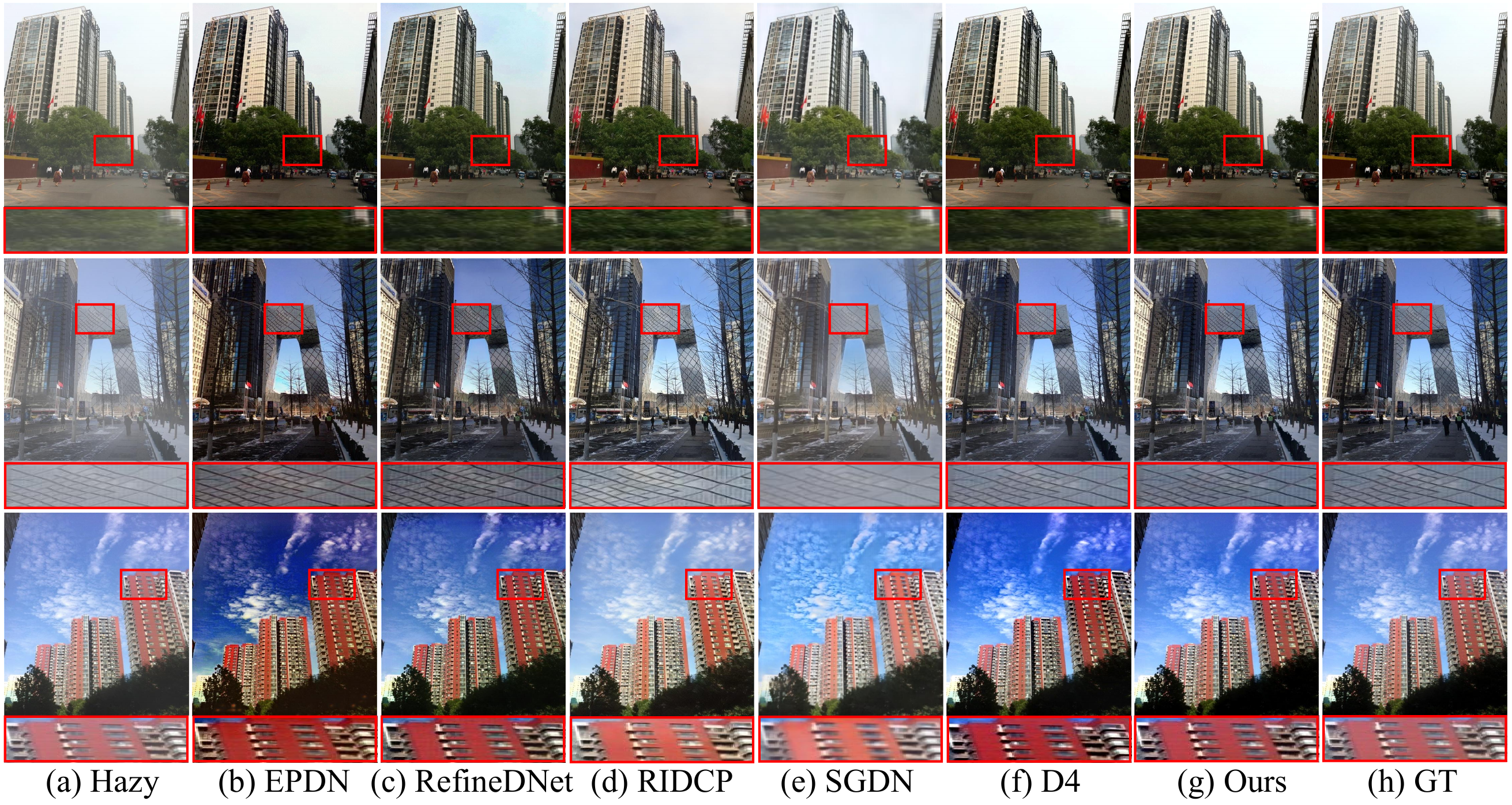}
\caption{Visual comparison of haze removal on samples from the SOTS-Outdoor datasets.}
\label{fig:ots}
\end{figure}
\begin{figure}
\centering
\includegraphics[width=\linewidth]{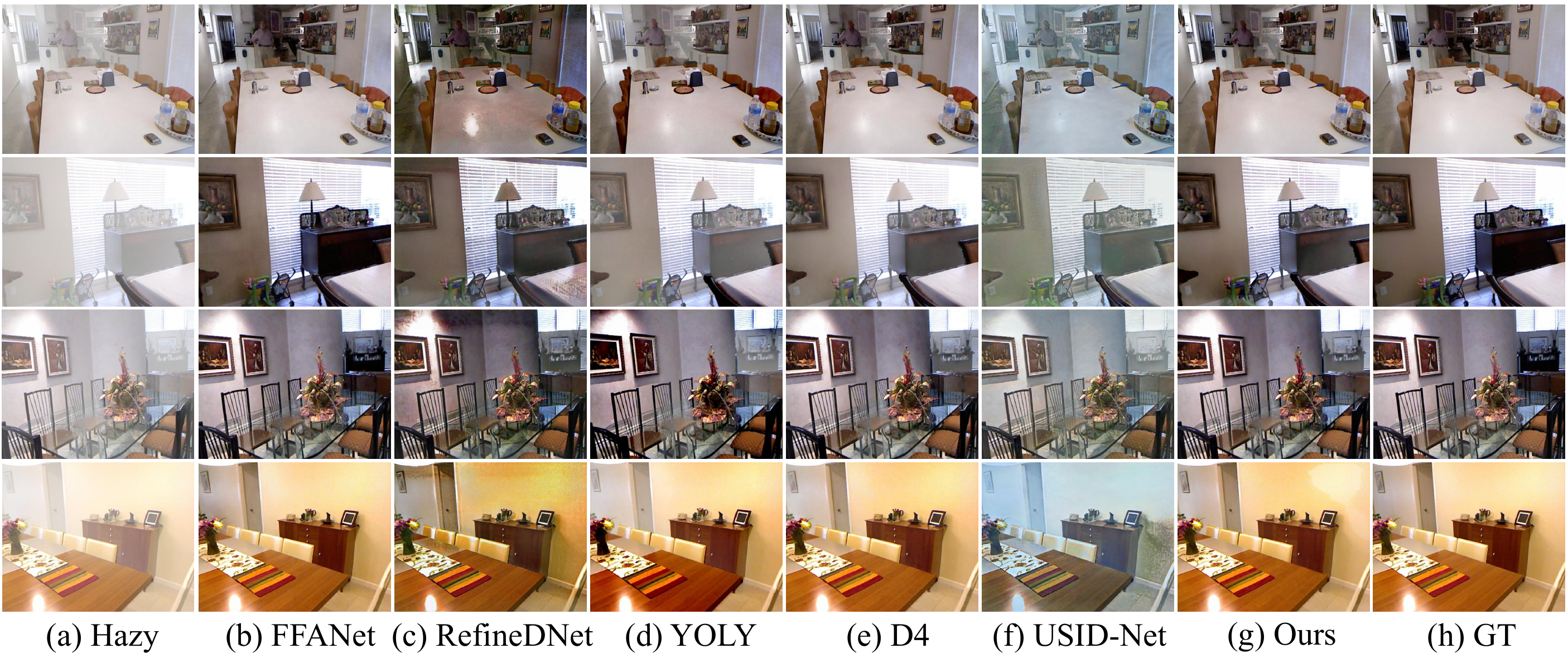}
\caption{Visual comparison of haze removal on samples from the SOTS-Indoor datasets.}
\label{fig:its}
\end{figure}
\begin{figure}
\centering
\includegraphics[width=\linewidth]{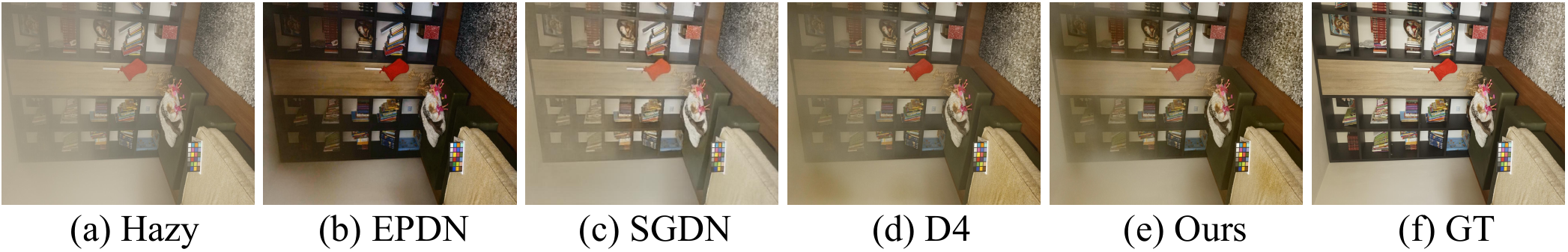}
\caption{Visual comparison of haze removal on samples from the I-HAZE datasets.}
\label{fig:ihaze}
\end{figure}
\begin{figure}
\centering
\includegraphics[width=\linewidth]{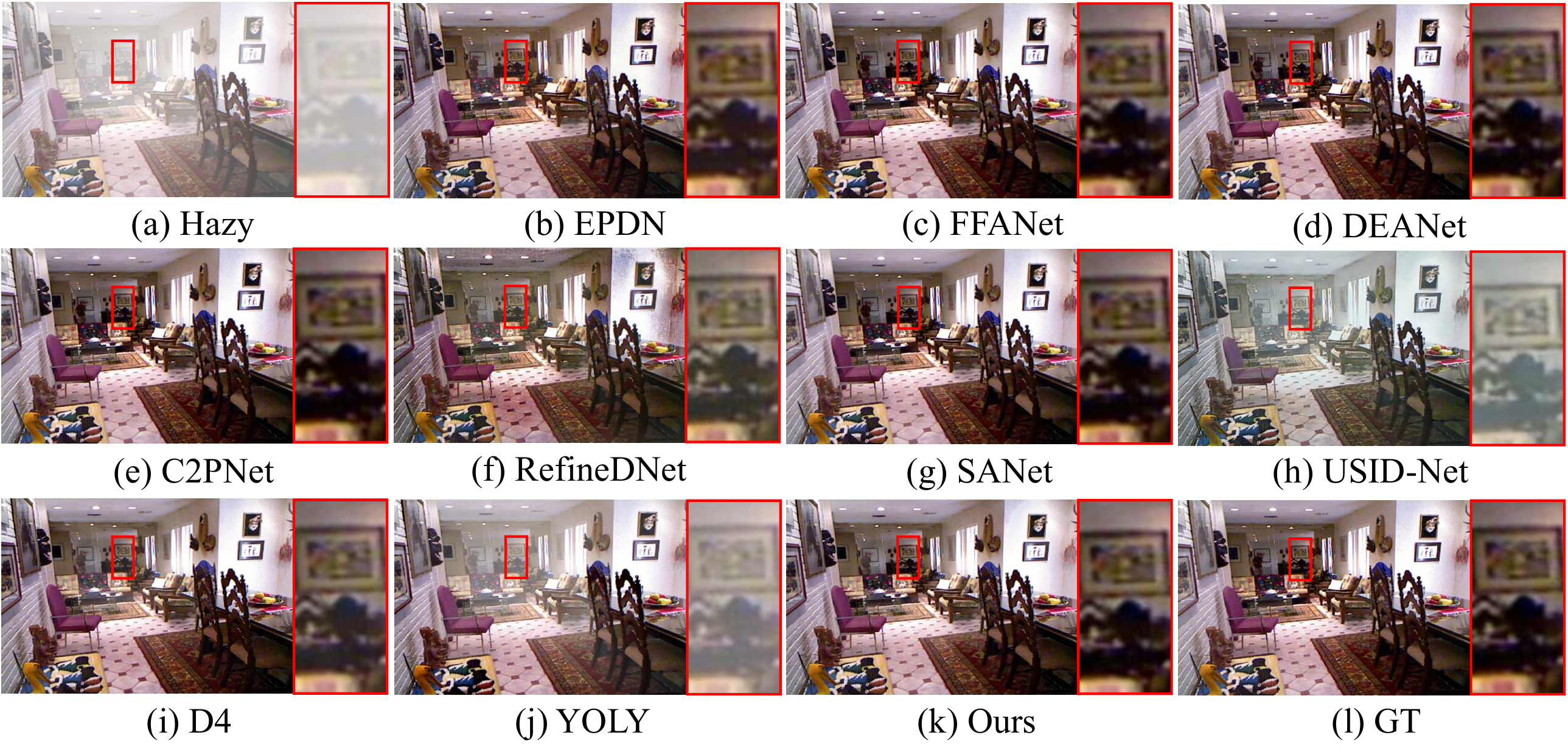}
\caption{Visual comparison of haze removal on samples from the SOTS-Indoor datasets. Compared with other dehazing methods, our method recovers more clearly.}
\label{fig:its_1}
\end{figure}
\begin{figure}
\centering
\includegraphics[width=\linewidth]{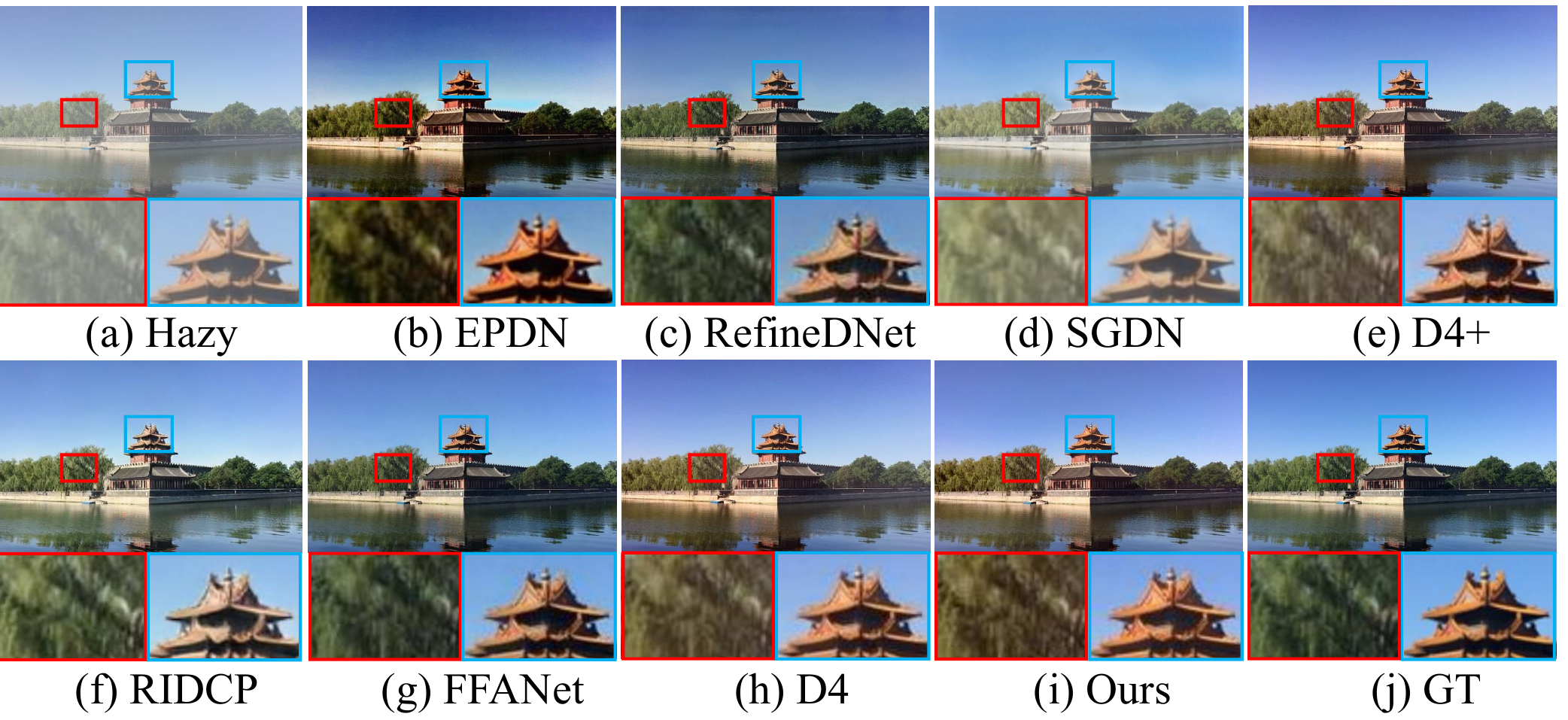}
\caption{Visual comparison of haze removal on samples from the HSTS datasets.}
\label{fig:hsts}
\end{figure}
\subsection{Experimental Setup}
\textbf{Training and Testing Data.} To train the model, we used the RESIDE \cite{li2018benchmarking} dataset. The indoor training set (ITS) of RESIDE contains 13,990 synthetic hazy images and 1,399 clean images, while the outdoor training set (OTS) contains 17,500 synthetic hazy images and 500 clean images. To be fairer, we also trained the baseline on the outdoor training set (OTS). In addition, we evaluate the proposed method on both synthetic and real-world datasets. The testing dataset includes a synthetic dataset with 500 indoor and 500 outdoor images, as well as real-world datasets such as the I-HAZE \cite{ancuti2018haze} dataset with 35 images, the PhoneHazy \cite{fan2023non} dataset with 40 images, the HSTS dataset with 10 images, the NH-Haze dataset with 5 images, the O-HAZE dataset with 45 images.\par
\textbf{Evaluation Metrics}. For quantitative comparison, we selected three metrics: PSNR, SSIM \cite{wang2004image}, and LPIPS \cite{zhang2018unreasonable}. These metrics are commonly used to evaluate the effectiveness of dehazing networks. Higher PSNR and SSIM values indicate better visual quality, while lower LPIPS values correspond to better visual quality.\par
\textbf{Experimental Details}. During the experiment, we carefully set up the training procedure for the proposed network model. Specifically, we used the Adam optimizer \cite{kingma2014adam}, with hyperparameters $ \beta_1 = 0.9 $, $ \beta_2 = 0.999 $, and a batch size of 4. Additionally, the learning rate was set to 0.0001. The model was trained using the PyTorch framework on an NVIDIA 3090 GPU. Through experimental validation, we found that after 1560,000 iterations, the network model achieved optimal performance.
\begin{figure}
\centering
\includegraphics[width=\linewidth]{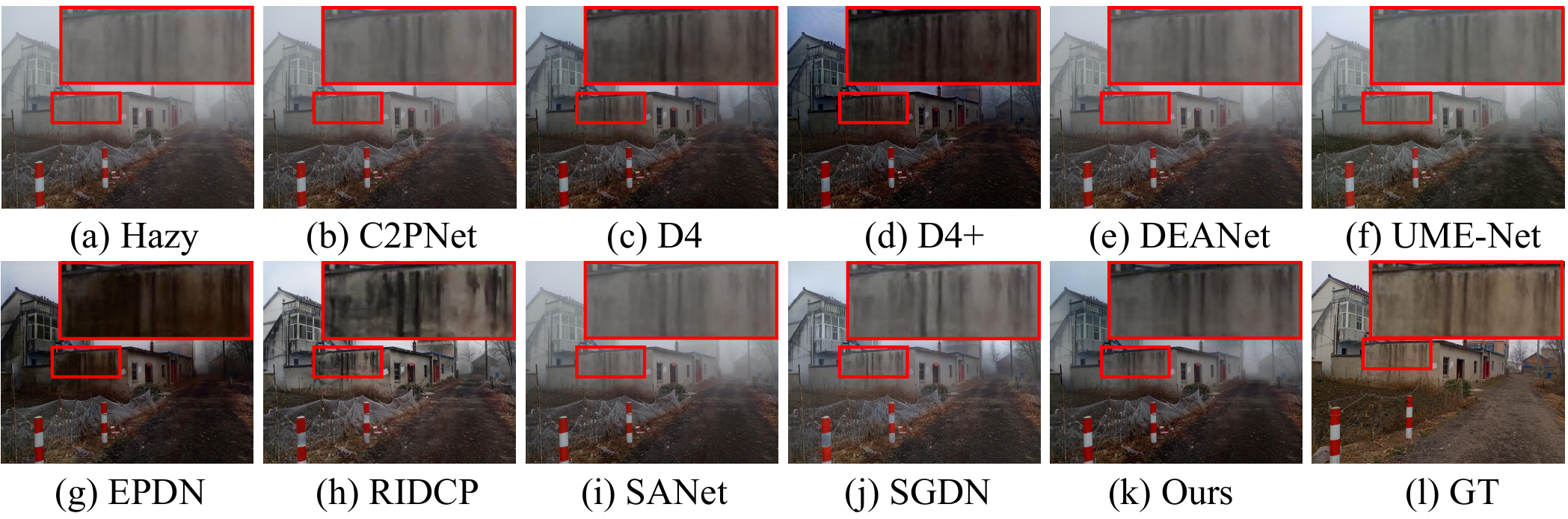}
\caption{Visual comparison of haze removal on samples from the PhoneHazy datasets.}
\label{fig:phonehazy}
\end{figure}
\begin{figure}
\centering
\includegraphics[width=\linewidth]{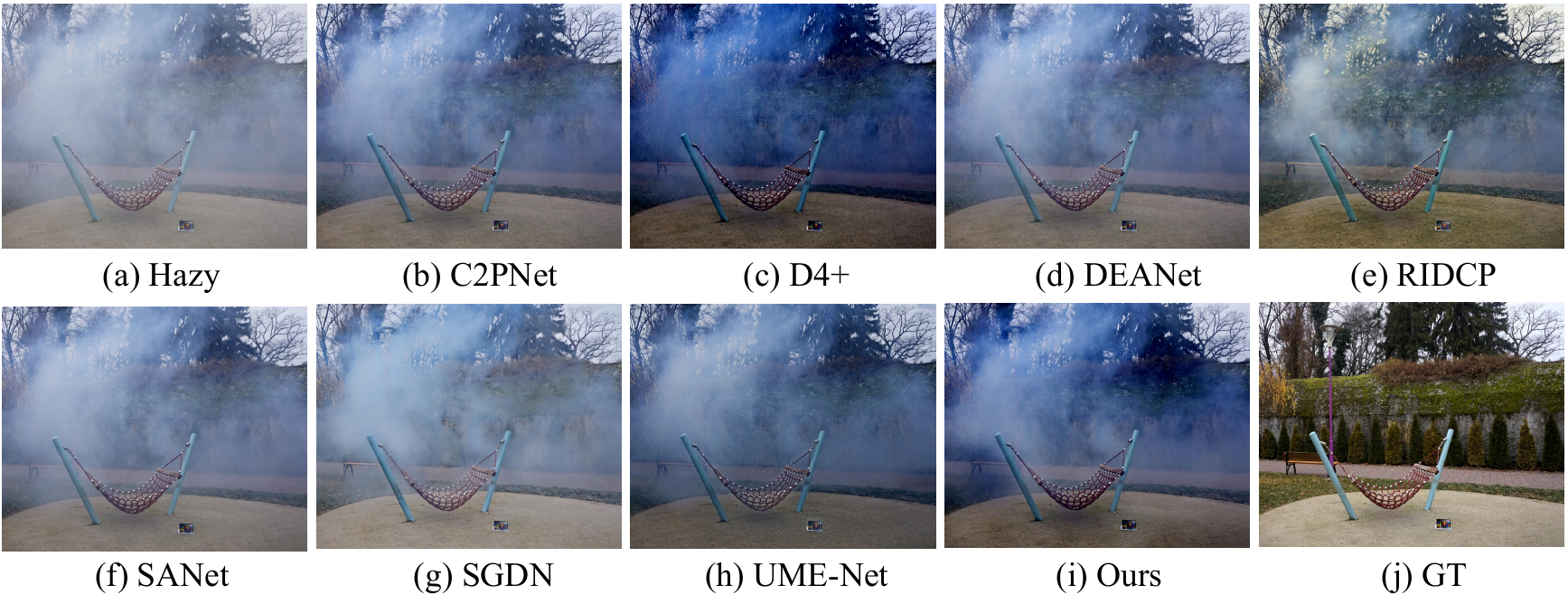}
\caption{Visual comparison of haze removal on samples from the NH-Haze datasets.}
\label{fig:nhhaze}
\end{figure}
\begin{table}[]
\caption{Quantitative comparisons (average PSNR/SSIM/LPIPS) with dehazing approaches on the SOTS-Outdoor and HSTS datasets. The first and second performances are shown in {\color{red} red} and {\color{blue} blue}, respectively.}
\scalebox{0.68}{
\centering
\begin{tabular}{lccccclccc}
\hline
\multicolumn{2}{l}{\multirow{2}{*}{\textbf{Method}}} & \multicolumn{1}{l}{} & \multicolumn{1}{l}{\textbf{Outdoor}} & \multicolumn{1}{l}{} & \multicolumn{1}{l}{} &  & \multicolumn{1}{l}{\textbf{HSTS}} & \multicolumn{1}{l}{} & \multicolumn{1}{l}{} \\ \cline{3-10} 
\multicolumn{2}{l}{}                        & Venue\&Year & PSNR$\uparrow$               & SSIM$\uparrow$        & LPIPS$\downarrow$       &  & PSNR$\uparrow$            & SSIM$\uparrow$        & LPIPS$\downarrow$       \\ \hline
\multirow{7}{*}{Paired}      & EPDN \cite{qu2019enhanced}         & CVPR 2019            & 20.30               & 0.887       & 0.104       &  & 21.30            & 0.867                & 0.112                \\
                             & FFANet \cite{qin2020ffa}       & AAAI 2020            & 25.14                       & 0.933                & \textcolor{blue}{0.036}                &  & 30.13                   & 0.939                & 0.036                \\
                             & SANet \cite{cui2023strip}        & IJCAI 2023           & \textcolor{blue}{32.71}                       & \textcolor{blue}{0.952}                & 0.037                &  & \textcolor{red}{32.57}                    & \textcolor{blue}{0.945}                & 0.033                \\
                             & C2PNet \cite{zheng2023curricular}       & CVPR 2023            & 31.06                       & 0.950                 & \textcolor{blue}{0.036}                &  & 31.12                    & 0.943                & \textcolor{blue}{0.031}                \\
                             & RIDCP \cite{wu2023ridcp}        & CVPR 2023            & 21.71                       & 0.850                 & 0.104                &  & 22.77                  & 0.854               & 0.087               \\
                             & DEANet \cite{chen2024dea}      & TIP 2024             & \textcolor{red}{35.25}                       & \textcolor{red}{0.985}                & \textcolor{red}{0.011}                &  & \textcolor{blue}{32.55}                    & \textcolor{red}{0.946}                & \textcolor{red}{0.030}                \\
                             & SGDN \cite{fang2024guided}         & AAAI 2025            & 18.29                       & 0.835                & 0.203                &  & 16.85                   & 0.806               & 0.211               \\ \hline
\multirow{10}{*}{Unpaired}   & RefineDNet \cite{zhao2021refinednet}   & TIP 2021             & 20.87                       & 0.906                & 0.088                &  & 20.94                    & 0.864                & 0.116                \\
                             & YOLY \cite{li2021you}         & IJCV 2021            & 16.41                       & 0.523                & 0.183                &  & 16.74                    & 0.483                & 0.202                \\
                             & USID-NET \cite{li2022usid}     & TMM 2022             & 24.21                       & 0.897                & 0.112                &  & 22.98                    & 0.754                & 0.137                \\
                             & D4 \cite{yang2022self}           & CVPR 2022            & 26.38                       & \textcolor{blue}{0.961}                & \textcolor{blue}{0.025}                &  & 26.10                    & 0.915                & \textcolor{red}{0.047}                \\
                             & *POGAN \cite{qiao2023learning}      & TCSVT 2023           & 25.90                       & 0.956                & /                    &  & 24.11                    & \textcolor{blue}{0.917}                & /                    \\
                             & Liu et al. \cite{liu2023reference}   & TCSVT 2023           & 23.22                       & 0.865                & /                    &  & 25.21                    & 0.882                & /                    \\
                             & *UCL \cite{wang2024ucl}        & TIP 2024             & 25.21                       & 0.927                & /                    &  & /                        & /                    & /                    \\
                            & *DFP-Net \cite{liu2024dfp}        & EAAI 2024             & 25.48                       & 0.907                & /                    &  & /                        & /                    & /                    \\
                             & *RPC-Dehaze \cite{lin2024toward}  & TCSVT 2024           & \textcolor{blue}{26.41}                       & \textcolor{red}{0.963}                & /                    &  & 25.76                    & \textcolor{red}{0.958}                & /                    \\
                             & *ODCR \cite{wang2024odcr}        & CVPR 2024            & 26.16                       & 0.960                & /                    &  & /                        & /                    & /                    \\
                             & \textbf{Ours}          & /                    & \textcolor{red}{26.98}                       & \textcolor{red}{0.963}                & \textcolor{red}{0.023}                &  & \textcolor{red}{26.11}                          & 0.915                     & \textcolor{blue}{0.049}                     \\ \hline
\end{tabular}
}
\label{table:outdoor_hsts}
\end{table}
\begin{table}[]
\caption{Quantitative comparisons (average PSNR/SSIM/LPIPS) with dehazing approaches on the PhoneHazy and NH-Haze datasets. The first and second performances are shown in {\color{red} red} and {\color{blue} blue}, respectively.}
\scalebox{0.68}{
\centering
\begin{tabular}{lcccccccc}
\hline
\multicolumn{2}{l}{\multirow{2}{*}{\textbf{Method}}} & \multicolumn{1}{l}{} & \multicolumn{3}{l}{\textbf{PhoneHazy}} & \multicolumn{3}{l}{\textbf{NH-Haze}}                               \\ \cline{3-9} 
\multicolumn{2}{l}{}                                 & Venue\&Year          & PSNR$\uparrow$        & SSIM$\uparrow$        & LPIPS$\downarrow$      & PSNR$\uparrow$                 & SSIM$\uparrow$                 & LPIPS$\downarrow$                \\ \hline 
\multirow{6}{*}{Paired}            & EPDN \cite{qu2019enhanced}            & CVPR 2019            & 13.47       & 0.468       & 0.559      & \textcolor{red}{14.25}                & \textcolor{red}{0.533}                & \textcolor{blue}{0.478}                \\
                                   
                                   & C2PNet \cite{zheng2023curricular}          & CVPR 2023            & \textcolor{blue}{17.98}       & \textcolor{red}{0.604}       & \textcolor{blue}{0.497}      & 12.33                & 0.468                & 0.549                \\
                                   & SANet \cite{cui2023strip}           & IJCAI 2023           & \textcolor{red}{18.08}       & \textcolor{red}{0.604}       & \textcolor{red}{0.496}      & 12.03                & 0.456                & 0.561                \\
                                   & RIDCP \cite{wu2023ridcp}           & CVPR 2023            & 14.96       & 0.478       & 0.530      & 12.27                & \textcolor{blue}{0.520}                & \textcolor{red}{0.474}                \\
                                   & DEANet \cite{chen2024dea}          & TIP 2024             & 17.35       & \textcolor{blue}{0.594}       & 0.501      & \textcolor{blue}{12.55}                & 0.481                & 0.538                \\

                                   & SGDN \cite{fang2024guided}            & AAAI 2025            & 16.29       & 0.579       & 0.539      & 12.45                & 0.392                & 0.692                \\ \hline
\multirow{4}{*}{Unpaired}          & D4 \cite{yang2022self}              & CVPR 2022            & 14.39       & 0.525       & 0.536      & \textcolor{blue}{13.20}                & \textcolor{red}{0.489}                & \textcolor{red}{0.526}                \\
                                   
                                   & D4+ \cite{yang2024robust}             & IJCV 2024            & 11.66       & 0.423       & \textcolor{blue}{0.526}      & 12.65                & 0.460                & 0.523                \\
                                   & UME-NET \cite{sun2024unsupervised}         & PR 2024              & \textcolor{red}{16.38}       & \textcolor{red}{0.566}       & \textcolor{red}{0.520}      & 11.33                & 0.410                & 0.577                \\
                                   & \textbf{Ours}            & /                    & \textcolor{blue}{15.76}            & \textcolor{blue}{0.558}            & 0.528           & \textcolor{red}{13.36} & \textcolor{blue}{0.486} & \textcolor{blue}{0.530} \\ \hline
\end{tabular}
}
\label{table:phonehazy_nhhaze}
\end{table}

\begin{table}[]
\caption{Quantitative comparisons (average PSNR/SSIM) with dehazing approaches on the SOTS-Indoor datasets.}
\centering
\scalebox{0.68}{
\begin{tabular}{c|ccccccccc}
\hline
              &CDD-GAN   &DCP  & RefineDNet & YOLY & USID-NET & D4 & *POGAN & *ADC-Net & \textbf{Ours} \\ \hline
\textbf{PSNR$\uparrow$}&24.61 &13.10   & 20.48         & 15.27         & 17.42          & 25.42          & 25.51          &25.52    & \textbf{25.73}      \\
\textbf{SSIM$\uparrow$} &0.918   &0.699    & 0.859         & 0.534         & 0.813          & 0.930          & 0.934          & 0.935    & \textbf{0.937}     \\ \hline
\end{tabular}
}
\label{table:indoor}
\end{table}
\begin{table}[]
\caption{Quantitative comparisons (average PSNR/SSIM) with dehazing approaches on the I-HAZE datasets.}
\centering
\scalebox{0.65}{
\begin{tabular}{c|cccccccc}
\hline
             &DCP & RefineDNet & D4 & *Cycle-SNSPGAN & *IC-Dehazing   & *RPC-Dehaze &UR2P-Dehaze  & \textbf{Ours} \\ \hline
\textbf{PSNR$\uparrow$} &13.10   & 15.50         & 15.71         & 15.36           & 15.70                        & 16.26 &16.36  &\textbf{16.47}    \\
\textbf{SSIM$\uparrow$} &0.699   & 0.710          & 0.753          & 0.740          & 0.726                        & \textbf{0.783} &0.770  &0.782   \\ \hline
\end{tabular}
}
\label{table:ihaze}
\end{table}
\begin{table}[h]
\caption{Comparison of the efficiency of different dehazing methods.}
\centering
\scalebox{0.78}{
\begin{tabular}{cccccc}
\hline
\multicolumn{2}{l}{\multirow{2}{*}{\textbf{Method}}}     & \multicolumn{1}{l}{\textbf{Efficiency}} & \textbf{}          & \textbf{}            & \textbf{}             \\ \cline{3-6} 
\multicolumn{2}{l}{}                            & Venue\&Year                    & Params(M) & FLOPs(GMac) & Runtime (ms) \\ \hline
\multirow{7}{*}{\textbf{Paired}}   & EPDN \cite{qu2019enhanced}       & CVPR 2019                               & 17.38              & 4.83                 & 41.86                 \\
                                   & FFANet \cite{qin2020ffa}    & AAAI 2020                               & 4.46               & 288.30                & 147.85                \\
                                   & PSD \cite{chen2021psd}        & CVPR 2021                               & 33.11              & 182.50                & 62.19                 \\
                                   & SANet \cite{cui2023strip}      & IJCAI 2023                              & 3.81               & 37.32                & 40.68                 \\
                                   & C2PNet \cite{zheng2023curricular}     & CVPR 2023                               & 7.17               & 463.49               & 305.77                \\
                                   & DEANet \cite{chen2024dea}    & TIP 2024                                & 3.65               & 34.19                & 61.29                 \\
                                   & SGDN \cite{fang2024guided}      & AAAI 2025                               & 13.32              & 53.40                & 149.76                \\ \hline
\multirow{5}{*}{\textbf{Unpaired}} 
                                   & RefineDNet \cite{zhao2021refinednet} & TIP 2021                                & 65.80               & 75.41                & 95.88                 \\
                                  & YOLY \cite{li2021you} & IJCV 2021                                & 32.00               & -                & -                 \\
                                   & D4 \cite{yang2022self}         & CVPR 2022                               & 10.70              & 2.25                 & 0.09                  \\
                                   & UME-Net \cite{sun2024unsupervised}    & PR 2024                                 & 59.84              & 189.64               & 270.81                \\
                                    & D4+ \cite{yang2024robust}    & IJCV 2024                                 & 10.70              & 2.25               & 0.08                \\
                                   & UR2P-Dehaze \cite{xue2026ur2p}    & PR 2025                                 & 10.75              & 2.29               & 0.14                \\
                                   & \textbf{Ours}       & /                                       & 13.38              & 16.72                & 0.11                  \\ \hline
\end{tabular}
}
\label{table:efficiency}
\end{table}
\subsection{Performance Evaluation}
To provide a comprehensive evaluation of our dehazing model, we conduct a comparative study against state-of-the-art dehazing methods. This comparison encompasses both qualitative and quantitative analyses.\par
\textbf{Qualitative Analysis}. Fig.~\ref{fig:ots}--\ref{fig:nhhaze} presents the results of the qualitative evaluation of different dehazing methods on several datasets,  including SOTS-Indoor, SOTS-Outdoor, HSTS, O-HAZE, NH-Haze, PhoneHazy, and I-HAZE. It can be observed that our method consistently achieves superior PSNR and SSIM scores in most scenarios. In particular, it demonstrates stronger generalization ability and detail restoration performance on complex real-world datasets such as PhoneHazy, O-HAZE and I-HAZE, which contain realistic haze degradation.\par
Further analysis of the visual results, as shown in Fig.~\ref{fig:ots}, reveals that our method outperforms competing approaches in restoring image structure, texture details, and color naturalness. Deep learning-based methods (e.g., RIDCP \cite{wu2023ridcp}, SGDN \cite{fang2024guided}, D4 \cite{yang2022self}, in terms of dehazing performance, often appear inadequate. In contrast, our proposed method, while maintaining overall image clarity, is capable of accurately recovering the structures of distant objects, removing residual haze, and avoiding over-enhancement. As shown in Fig.~\ref{fig:its}, USID-NET \cite{li2022usid} and RefineDNet \cite{zhao2021refinednet} methods tend to exhibit significant deviations in color recovery. This results in greater visual consistency and practical applicability of the images.\par
\textbf{Quantitative Analysis}. In order to comprehensively evaluate the performance of our proposed dehazing method, we conducted quantitative analyses with multiple dehazing methods, such as SANet \cite{cui2023strip}, FFANet \cite{qin2020ffa}, IC-Dehazing \cite{gui2023illumination}, RPC-Dehaze \cite{lin2024toward}, EPDN \cite{qu2019enhanced}, PSD \cite{chen2021psd}, Cycle-SNSPGAN \cite{wang2022cycle}, CDD-GAN \cite{chen2022unpaired}, D4 \cite{yang2022self}, POGAN \cite{qiao2023learning}, C2PNet \cite{zheng2023curricular}, DCP \cite{he2010single}, ADC-Net \cite{wei2024robust}, RIDCP \cite{wu2023ridcp}, UME-Net \cite{sun2024unsupervised}, SGDN \cite{fang2024guided}, ODCR \cite{wang2024odcr}, UCL \cite{wang2024ucl}, EENet \cite{cui2025eenet} and DEANet \cite{chen2024dea}, on multiple publicly available datasets. For example, we compare our method with current state-of-the-art dehazing methods using standard evaluation metrics such as peak signal-to-noise ratio (PSNR), structural similarity index (SSIM), and Learned Perceptual Image Patch Similarity(LPIPS). \autoref{table:outdoor_hsts}--\ref{table:ihaze} Experimental results show that our method achieves excellent performance on all test datasets. Specifically, on the SOTS-Indoor dataset, our method achieves a PSNR of 25.73 dB and an SSIM of 0.937, outperforming existing methods. On the SOTS-Outdoor dataset, our method achieves a PSNR of 26.98 dB and an SSIM of 0.963, which demonstrates the strong dehazing capability of our method in outdoor scenes. The I-HAZE dataset, which consists of real haze and corresponding haze-free indoor images, also achieves PSNR and SSIM values of 16.47 and 0.783 dB, respectively, which are ahead of existing methods. More experimental data are shown in detail in the corresponding tables. The results confirm that our method achieves satisfactory performance in real-world applications. The “ * ” in front of the method name indicates that the code is not open source, and the results are in the cited paper. \par
\begin{table}[]
\caption{Quantitative comparisons (average PSNR/SSIM) with dehazing approaches on the O-HAZE datasets.}
\centering
\scalebox{0.78}{
\begin{tabular}{c|ccccccccc}
\hline
              & SANet & RIDCP & C2PNet & DEANet & SGDN & D4 & UME-Net &EENet  & \textbf{Ours} \\ \hline
\textbf{PSNR$\uparrow$} &14.66    &15.97     &15.26      &15.49       &14.86      &16.65    &15.65    &14.81 &\textbf{16.77}   \\
\textbf{SSIM$\uparrow$} &0.605  &0.649    & 0.620  &0.622   & 0.471    &0.666  &0.627  &0.608  &\textbf{0.676}  \\ \hline
\end{tabular}
}
\label{table:indoor}
\end{table}
As shown in~\autoref{table:efficiency}, our dehazing model, which does not require paired training data, demonstrates outstanding efficiency with significantly fewer parameters and extremely low FLOPs. Although lightweight models such as SANet \cite{cui2023strip}, FFANet \cite{qin2020ffa}, and DEANet \cite{chen2024dea} also exhibit competitive performance in terms of parameter count and computational cost, they rely on supervised training and fall short in dehazing performance compared to our method. The experimental results further demonstrate that our approach outperforms existing methods in both efficiency and dehazing quality, while also exhibiting greater stability.
\begin{table}[ht!]
\caption{Ablation experiments on indoor and outdoor datasets. `` \ding{55} "(resp. `` \checkmark~") means the corresponding module is unused (resp. used).}
\centering
\scalebox{0.95}{
\begin{tabular}{ccccccccc}
\hline
\multicolumn{1}{l}{\textbf{Type}} & \multicolumn{3}{l}{\textbf{Method}} & \multicolumn{1}{l}{} & \multicolumn{4}{l}{\textbf{Outdoor}}      \\ \cline{2-4} \cline{6-9} 
          & KAN-CID        & IDRM       & DREM       &                      & PSNR$\uparrow$  & SSIM$\uparrow$  & LPIPS$\downarrow$ & FID$\downarrow$    \\ \hline
          V1 & \ding{55}          & \ding{55}         & \ding{55}          &                      & 26.38 & 0.961 & 0.025 & 7.084      \\
          V2 & \checkmark          & \ding{55}         & \checkmark          &                      & 26.60 & 0.959 & 0.024 & 7.253       \\
          V3 & \checkmark          & \checkmark         & \ding{55}          &                      & 26.80 & 0.961 & 0.024 & 6.574      \\
          V4 & \ding{55}          & \checkmark         & \checkmark         &                      & 26.71  &0.961  &0.025  &6.675       \\ \hline
          V5 & \checkmark          & \checkmark        & \checkmark          &                      & \textbf{26.98} & \textbf{0.963} & \textbf{0.023} & \textbf{6.335}      \\ \hline
\end{tabular}
}
\label{table:ablation_ots}
\end{table}
\begin{figure}
\centering
\includegraphics[width=\linewidth]{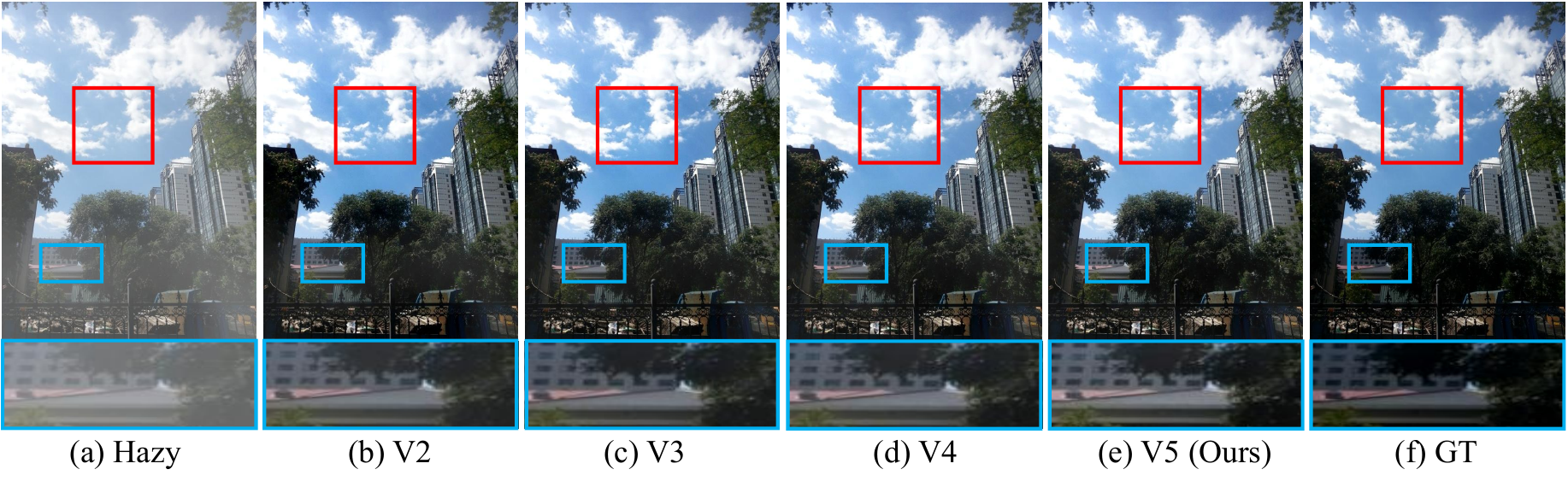}
\caption{Visual comparison results of ablation experiments.}
\label{fig:ablation}
\end{figure}
\subsection{Ablation Study}
To evaluate the effectiveness of our approach in image decontamination tasks, we conducted ablation experiments on different modules using the SOTS-Outdoor test set. As shown in~\autoref{table:ablation_ots}, specifically, the integration of Variant 2 (V2): KAN-CID greatly enhances the ability of the model to learn the nonlinear degradation relation between gray-scale images and clean images, thus bringing the restored images closer to the distribution of the ground truth. In order to achieve an effective balance between degradation normalization and content fidelity, Variant 3 (V3): further shows that our strategy focuses on implicit neural representation and irregular degradation features. This improves the representational capacity. and generalization performance of the boosted model. \par
\begin{table}[]
\caption{Ablation study on different numbers of training iterations (in units of ten thousand).}
\centering
\scalebox{1}{
\begin{tabular}{c|cccccc}
\hline
              & \textbf{N=40} & \textbf{N=70} & \textbf{N=100} & \textbf{N=130} & \textbf{N=156} & \textbf{N=160} \\ \hline
\textbf{PSNR$\uparrow$} & 25.63         & 26.39         & 25.90          & 26.47          & \textbf{26.98}          & 26.27          \\
\textbf{SSIM$\uparrow$} & 0.953         & 0.960         & 0.960          & 0.959          & \textbf{0.963}          & 0.962          \\ \hline
\end{tabular}
}
\label{table:iterations}
\end{table}
To investigate the effect of training iterations on model performance, we conduct an ablation study with different iteration counts. As shown in~\autoref{table:iterations}, both PSNR and SSIM exhibit an overall increasing trend as the number of iterations grows, indicating progressive improvements in reconstruction quality. The model achieves its best performance at $N = 156$, reaching a PSNR of 26.98 and an SSIM of 0.963. Notably, a slight performance drop is observed at $N = 160$, which may be attributed to overfitting or training saturation. These findings highlight the importance of carefully selecting the number of training iterations to ensure a balance between convergence and generalization.\par
As illustrated in Fig.~\ref{fig:ablation}, Variant 2 (V2): represents the visual results without the IDRM module, Variant 3 (V3): represents the results without the HFDM module, Variant 4 (V4): represents the results without the KAN-CID module, and Variant 5 (V5): represents the dehazing visual results of our complete dehazing model. The results demonstrate that each of these modules significantly enhances the visual quality of the dehazed images.  Additional ablation studies are presented in~\autoref{table:ablation_ots}.
\section{Conclusion}\label{Conclusion}
In this paper, we propose a novel unpaired image dehaze method, called NeDR-Dehaze, which can achieve competitive visual results on multiple publicly available datasets. Firstly, to balance between fine-grained feature representation in inhomogeneous hazy distribution and global consistency modeling. We design a channel-independent and channel-dependent mechanism, KAN-CID, which effectively enhances the model's ability to capture spatial variations and non-local structural features of haze distribution in complex scenes. Then, to alleviate the model's dependence on explicit feature extraction and physical modeling, we design the implicit dense residual module, which enables the features learned by the model to better contribute to haze removal and improves the robustness of the model in complex scenes. Experimental results show that the proposed method achieves excellent dehazing performance on a variety of public and real-world haze datasets, resulting in more effective recovery of image details and textures.




\bibliographystyle{elsarticle-num} 


\bibliography{template.bib}
\end{document}